\pdfoutput=1
\documentclass{article}
\usepackage[nonatbib,preprint]{neurips_template} 

\usepackage{amsmath,amssymb,amsfonts}
\usepackage{algorithmic}
\usepackage{graphicx}
\usepackage{textcomp}
\usepackage{times}
\usepackage{soul}
\usepackage{url}
\usepackage{xcolor, colortbl}
\usepackage{adjustbox}
\usepackage{tabularx}
\usepackage{multirow}
\usepackage{array}
\usepackage{xspace}
\usepackage[normalem]{ulem}
\usepackage{wrapfig,lipsum,booktabs}
\usepackage{amsthm}

\usepackage[hidelinks]{hyperref}
\usepackage[utf8]{inputenc}
\usepackage[small]{caption}
\usepackage{booktabs}
\usepackage{algorithm}
\usepackage{algorithmic}
\usepackage{subfigure}
\usepackage{caption}
\usepackage{bm}

\newtheorem*{theorem*}{Theorem}

\newcommand{\makeappendixtitle}{%
    \vbox{%
    \hsize\textwidth
    \linewidth\hsize
    \vskip 0.1in
    \hrule height 4pt
    \vskip 0.25in
    \vskip -\parskip%
    \centering
    {\LARGE\bf {Appendix} \par}
    \vskip 0.29in
    \vskip -\parskip
    \hrule height 1pt
    \vskip 0.09in%
  }
}

\renewcommand{\mathbf}[1]{#1}
\renewcommand{\boldsymbol}[1]{#1}
\newtheorem{theorem}{Theorem}

\usepackage{amsmath,amsfonts,bm}

\def\eqref#1{equation~\ref{#1}}

\def\1{\bm{1}}

\DeclareMathAlphabet{\mathsfit}{\encodingdefault}{\sfdefault}{m}{sl}
\SetMathAlphabet{\mathsfit}{bold}{\encodingdefault}{\sfdefault}{bx}{n}

\newcounter{gaocomm} 
  
\definecolor{blue-violet}{rgb}{0.0, 0.81, 0.90}
\definecolor{mygreen}{rgb}{0.0, 0.5, 0.0}
\definecolor{awesome}{rgb}{1.0, 0.13, 0.32}
\definecolor{bostonuniversityred}{rgb}{0.8, 0.0, 0.0}

\usepackage[maxbibnames=99]{biblatex}  
\title{
Universal Deep GNNs: Rethinking Residual Connection in GNNs from a Path Decomposition Perspective for Preventing the Over-smoothing
}

\author{
Jie Chen$^{1}$, Weiqi Liu$^{1}$, Zhizhong Huang$^{1}$, Junbin Gao$^{2}$, Junping Zhang$^{1}$, Jian Pu$^{3}$ \\
$^{1}$Shanghai Key Lab of Intelligent Information Processing, School of Computer Science\\ Fudan University, 
Shanghai 200433, P. R. China \\ $^{2}$Discipline of Business Analytics, The University of Sydney Business School\\
The University of Sydney, Sydney, NSW 2006, Australia \\
$^{3}$Institute of Science and Technology for Brain-Inspired Intelligence\\
Fudan University, Shanghai 200433, P. R. China \\
$^{1,3}${\{chenj19, wqliu20, zzhuang19, jpzhang,jianpu\}@fudan.edu.cn, $^{2}$junbin.gao@sydney.edu.au}
}

\begin{document}
\maketitle
\begin{abstract}
The performance of GNNs degrades as they become deeper due to the over-smoothing. Among all the attempts to prevent over-smoothing,  residual connection is one of the promising methods due to its simplicity. However, recent studies have shown that GNNs with residual connections only slightly slow down the degeneration. The reason why residual connections fail in GNNs is still unknown. In this paper, we investigate the forward and backward behavior of GNNs with residual connections from a novel path decomposition perspective. We find that the recursive aggregation of the median length paths from the binomial distribution of residual connection paths dominates output representation, resulting in over-smoothing as GNNs go deeper. Entangled propagation and weight matrices cause gradient smoothing and prevent GNNs with residual connections from optimizing to the identity mapping. Based on these findings, we present a Universal Deep GNNs (UDGNN) framework with cold-start adaptive residual connections (DRIVE) and feedforward modules. Extensive experiments demonstrate the effectiveness of our method, which achieves state-of-the-art results over non-smooth heterophily datasets by simply stacking standard GNNs.
\end{abstract}
\section{Introduction}
In recent years, Graph Convolutional Networks (GCNs) have been successfully applied to a wide range of graph applications, including social networks~\cite{tnnls-compre}, traffic prediction~\cite{cui2019traffic}, knowledge graphs~\cite{park2019estimating}, drug reaction~\cite{do2019graph} and recommendation system~\cite{he2020lightgcn}.
The behavior of most GNNs is similar to a Laplacian smoother~\cite{li2018deeper} or low-pass filter~\cite{nt2019revisiting}, which learns node representation by recursively aggregating neighbor information.
Despite the remarkable success, such a recursive smoothing process of GNNs usually results in the over-smoothing problem, i.e., all node representations will converge to indistinguishable~\cite{huang2020tackling,li2018deeper,oono2019graph,yan2021two}.

The following two issues are crucial among the limitation of GNNs caused by over-smoothing.
First, the performance of GNNs usually degenerates when more graph convolution operations are applied~\cite{li2018deeper,liu2020towards,zhou2020towards}. As a result, recent GNNs often use shallow architectures (e.g., 2 or 3 stacked convolution operations)~\cite{Hamilton2017InductiveRL,GCN,GAT}, which ignore distant connections and can only exploit local structural information~\cite{abu2019mixhop,xu2018representation}. Second, in some heterophily graphs that contain non-smooth signals, i.e., many connected nodes belonging to different classes, the performance of both shallow and deep GNNs is usually inferior to that of simple MLPs without neighbor aggregation~\cite{bo2021beyond,pei2019geom, zhu2020beyond}.

\begin{figure}[t]
\centering
\subfigure[GraphConv with different skip connections]
{
\begin{minipage}{0.45\linewidth}
\centering
\includegraphics[width=\linewidth]{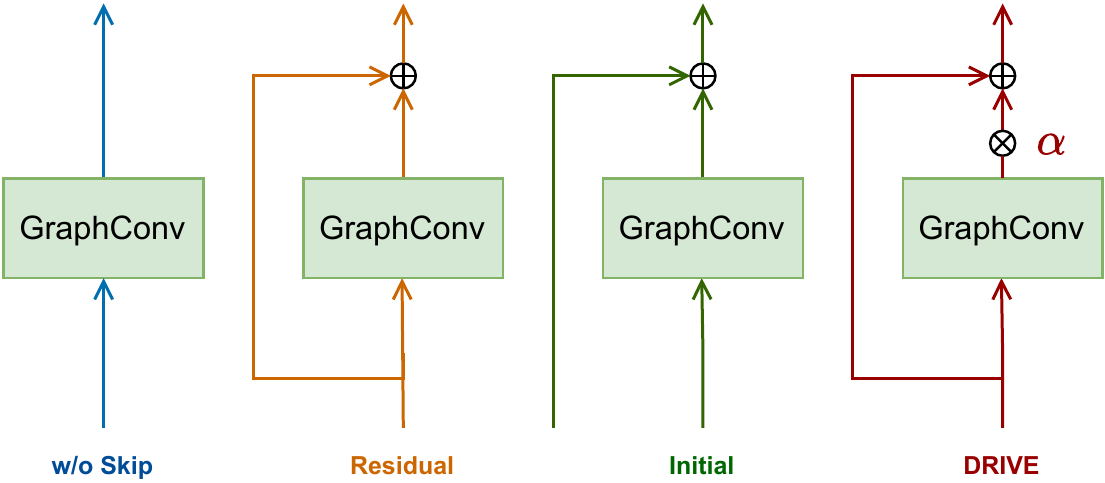}
\end{minipage}
}
\subfigure[Over-smoothing in Cora-Random]
{
\begin{minipage}{0.45\linewidth}
\centering
\includegraphics[width=\linewidth]{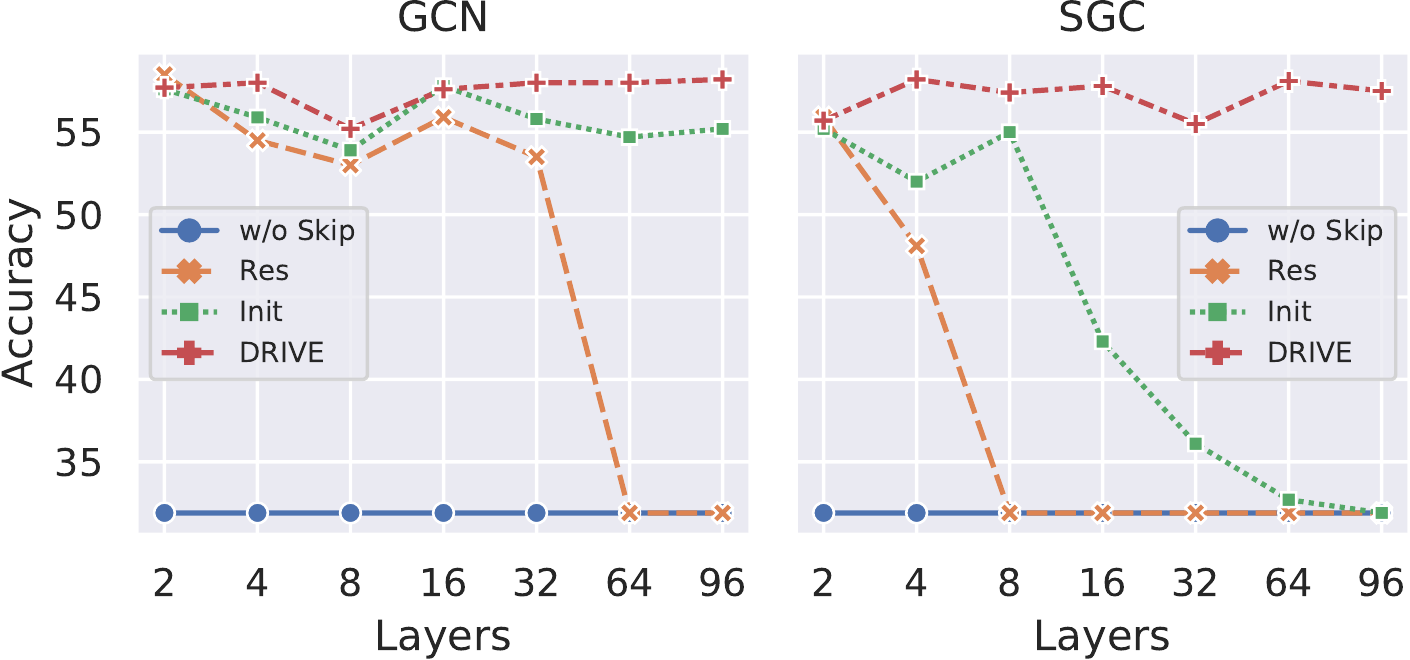}
\end{minipage}
}
\caption{Illustration of GNNs with different skip connections and the performance comparison on the fully-connected Cora-Random (noisy neighbor information) dataset in terms of GraphConv layers. Notice that although the skip connections are helpful for shallow networks, their performance degrades as more layers are stacked, except for the proposed DRIVE.
}
\label{fig:oversmooth}
\vspace{-0.5cm}
\end{figure}

One straightforward way to alleviate over-smoothing is to use the residual connection, as in ResNet~\cite{he2016deep,he2016identity}. 
Unfortunately, researchers found that simple residual connections only partially relieve the over-smoothing problem, and the performance of the model still degrades as more layers are stacked~\cite{chen2020simple,chen2022bag,GCN,oono2019graph,rong2019dropedge}, as shown in Figure~\ref{fig:oversmooth}. 
The convergence of GNNs with residual connections has been analyzed using lazy random walk~\cite{chen2020simple,xu2018representation} or the subspace perspective~\cite{oono2019graph}, and variants of JK connections~\cite{xu2018representation} and initial connections~\cite{chen2020simple} were proposed for training DeepGNNs.
However, as those convergence analysis of residual connections in GNNs ignores the optimization process of the weight matrix, their convergence conditions are usually not guaranteed in real applications~\cite{cong2021provable}, and thus it cannot be well explained why residual connections fail for GNNs. 

In this paper, 
we first present a novel forward and backward analysis of the stacking graph convolution layer with various kinds of skip connections, as shown in Figure~\ref{fig:oversmooth}(a), from a path decomposition perspective. 
This perspective views GNNs with skip connections as a collection of many propagation paths instead of a single deep network.
To explain why residual connections fail in DeepGNNs, we systematically analyze the behavior from both the forward path decomposition and backward gradient optimization. Then, we find that:
1) Due to the recursive aggregation process, the binomial distribution of residual connection paths causes the medium-length path to dominate the output representation, which results in over-smoothing when the GNNs become deeper.
2) The weight matrix in each graph convolution layer may provide a path selection ability. Compared with the SGC with residual connections, the ResGNNs can deactivate longer paths in the ideal scenario by adequately setting the weight matrix to learn identity mapping. 
3) However, unlike ResNet, the entanglement of the propagation matrix and weight matrix would cause the phenomenon of gradient smoothing, which would prevent the GNNs with residual connections from optimizing to identity mapping. 
Therefore, we speculate that the reason why the residual connections fail to learn identity mapping in DeepGNNs lies in the optimization process of the weight matrix.

Building on the aforementioned analysis from the path decomposition view, we propose a simple yet effective \textit{col{\bf D} sta{\bf R}t adapt{\bf IV}e r{\bf E}sidual} (\textbf{DRIVE}) connection for GNNs to prevent over-smoothing and effectively scale up the depth. It multiplies a trainable coefficient $\alpha$ to the graph convolution unit at each residual connection to control the smoothness.
Moreover, we impose the {\it direct pass} identity mapping for the entire network at the beginning by initializing $\alpha$ with 0. This would lead to favorable forward and backward propagation properties, i.e., prevent over-smoothing at initialization, which can solve the gradient smoothing problem and accelerate model training.
Furthermore, we propose a novel {\bf U}niversal {\bf D}eep {\bf GNN}s (\textbf{UDGNN}) framework that applies the DRIVE connection on both the GraphConvolution and the following FeedForward module. The UDGNN framework is GraphConv agnostic and can give the freedom of any standard GNNs to learn identity mapping or DeepMLP mapping and become deeper and more powerful.

\section{Related work}
\textbf{Oversmooth Analysis.} Recently, several works have attempted to understand and alleviate over-smoothing in GNNs~\cite{chen2020simple, cong2021provable,dong2021attention,huang2020tackling,liu2020towards, oono2019graph}.
The Markov process of simplified GNNs without weight matrices was studied in~\cite{chen2020simple,liu2020towards,xu2018representation}, which theorized that the stacking propagation matrix would converge to the eigenvectors associated with degree information. According to~\cite{chen2020simple}, the residual connection without weight matrices replicates a lazy random walk, converges to a stationary state, and leads to over-smoothing.
Furthermore, the subspace theorem for the graph convolution with weight matrix was proposed and showed that the node representation would exponentially converge to the subspace, and the residual connection only plays the role of slowing the convergence rate~\cite{huang2020tackling,oono2019graph}. Other works propose Dirichlet energy-based analysis~\cite{cai2020note,zhou2021dirichlet} and obtain a similar result to the subspace theorem. However, those analyses mainly focus on the forward behavior of GNNs and do not well explain the reason why the residual connection fails for DeepGNNs.

\noindent \textbf{Decouple GNNs.} 
Most literature reports that the speed of degeneration of GCN is faster than SGC~\cite{wu2019simplifying} without weight matrix and hypothesizes that the entanglement of transformation and propagation significantly compromises the performance of DeepGNNs~\cite{liu2020towards}. 
Hence, the decouple GNNs are proposed to separate the propagation and transformation~\cite{chien2020adaptive,cong2021provable,klicpera2018predict,liu2020towards}, allowing for broader propagation. However, they are not real deep models but only enhance the nodes' perceptual field by multiple hops. In addition, those designs are not easily compatible with standard graph convolutions.

\noindent \textbf{Skip connection in GNNs.}
Several works apply skip connections to relieve the over-smoothing issue, including: 1) residual connection with dilated-conv or message normalization~\cite{li2020deepergcn,li2021deepgcns}, 2) dense or JK connection to combine all previous layers' representations~\cite{xu2018representation,zhu2020beyond}, and 3) initial connection with the initial node embedding~\cite{bo2021beyond,chen2020simple,li2019predicting}. 
However, directly introducing plain residual connections without regularization does not work for DeepGNNs, and the dense-like connection requires large memory usage. Instead, the initial connection is widely used in the design of GNNs to alleviate over-smoothing.
Therefore, by investigating the difference in behavior between the initial and residual in a fundamental way, we propose a novel DRIVE connection for DeepGNNs without modifying the GraphConv kernel or applying normalization and regularization techniques designed for graphs.

\noindent \textbf{Regularization in GNNs.}
Several works propose training regularization techniques for graphs to alleviate over-smoothing. On the one hand, analogous to BatchNorm~\cite{Ioffe2015BatchNA}, the PairNorm~\cite{zhao2019pairnorm}, NodeNorm~\cite{Zhou2020EffectiveTS}, MessageNorm~\cite{li2020deepergcn}, and GroupNorm~\cite{zhou2020towards} adjust the statistics over graphs to slow down the over-smoothing. On the other hand, borrowing the idea of dropout~\cite{srivastava2014dropout}, DropEdge~\cite{rong2019dropedge}, DropNode~\cite{feng2020graph}, and SkipNode~\cite{lu2021skipnode} introduce the randomly dropped techniques into graphs to alleviate over-smoothing. 
\section{Preliminaries}
\subsection{Notation and Problem Setting}
Consider an undirected graph ${\cal{G} = (\cal{V}, \cal{E})}$, with $N$ nodes and $m$ edges. We use $\{1,\dots,N\}$ to denote the node index of $\cal{G}$, wheres $d_j$ denotes the degree of node $j$ in $\cal{G}$. Let ${\mathbf{A}\in R^{N\times N}}$ be the adjacency matrix and ${\mathbf{D}\in R^{N\times N}}$ be the diagonal degree matrix. 
Each node is given a $d$-dimensional feature representation $\mathbf{x}_i$ and a $c$-dimensional one-hot class label $\mathbf{y}_i$. The feature inputs are then formed by $\mathbf{X}=[\mathbf{x}_1, \cdots, \mathbf{x}_N]$, and the labels are $\mathbf{Y}=[\mathbf{y}_1, \cdots, \mathbf{y}_N]$.
Given the labels ${\mathbf{Y}}_{\mathcal{L}}$ of the nodes ${\mathcal{L}} \subset {\mathcal{V}}$, the task of node classification is to predict the labels $\mathbf{Y}_U$ of the unlabeled nodes ${\mathcal{U}} = {\mathcal{V}} \setminus {\mathcal{L}}$ by exploiting the graph adjacency matrix $A$ and the features $\mathbf{X}$ corresponding to all the nodes. In addition, given a set of node labels $Y$ over graphs, the notion of homophily and heterophily indicates the smoothness of the signal of the label. If connected nodes tend to have the same class, the graphs correspond to high homophily and low heterophily.

\subsection{Graph Neural Networks}
The general GNN is composed of information aggregation and feature transformation~\cite{Hamilton2017InductiveRL,CN,GAT,you2020design}. 
For the $l$-th layer of a GNN, we use $\mathbf{H}^{l}$ to represent the embedding of nodes and $\mathbf{H}^{0}$ to represent the initial feature $\mathbf{X}$ or a projection of $\mathbf{X}$ for dimension reduction. 
Then, the general $l$-th layer Graph Convolution can be formulated as
\begin{align}
\mathbf{H}^l &= \operatorname{GraphConv}(\mathbf{A},\mathbf{H}^{l-1})=\sigma(\mathbf{PH}^{l-1}\mathbf{W}^l),  
\end{align}
where the propagation matrix $\mathbf{P}$ achieves information aggregation from neighbors, and the weight matrix $\mathbf{W}$ completes feature transforms. The propagation matrix $\mathbf{P}$ is usually calculated based on the adjacent matrix ${\bf A}$ and degree matrix ${\bf D}$ as in GCN~\cite{GCN}, SGC~\cite{wu2019simplifying} and GraphSAGE~\cite{Hamilton2017InductiveRL} or the attention mechanism in GAT~\cite{GAT}. Moreover, various GNNs remove the nonlinear ReLU activation $\sigma$ and weight matrix $\mathbf{W}$ also achieve comparable performance~\cite{chien2020adaptive,he2020lightgcn,li2019predicting,wu2019simplifying}. Please refer to the review for more details~\cite{tnnls-compre}. 
\section{Analysis of DeepGNNs from the Path Decomposition View}
To better understand the role of skip connections in GNNs and explain why residual connections fail, 
we develop a first path decomposition perspective to study the forward and backward behavior of GNNs with different skip connections. In contrast to previous studies, we show that the GNNs with residual connections have more expressive capacity than initial connections in the forward analysis, and the weight matrix plays an important role. In the backward analysis, we find that the gradient smoothing problem causes the optimization difficulty and prevents the residual connection from learning the identity mapping to avoid over-smoothing. 
\begin{figure}[t]
\centering
\centerline{\includegraphics[width=\linewidth]{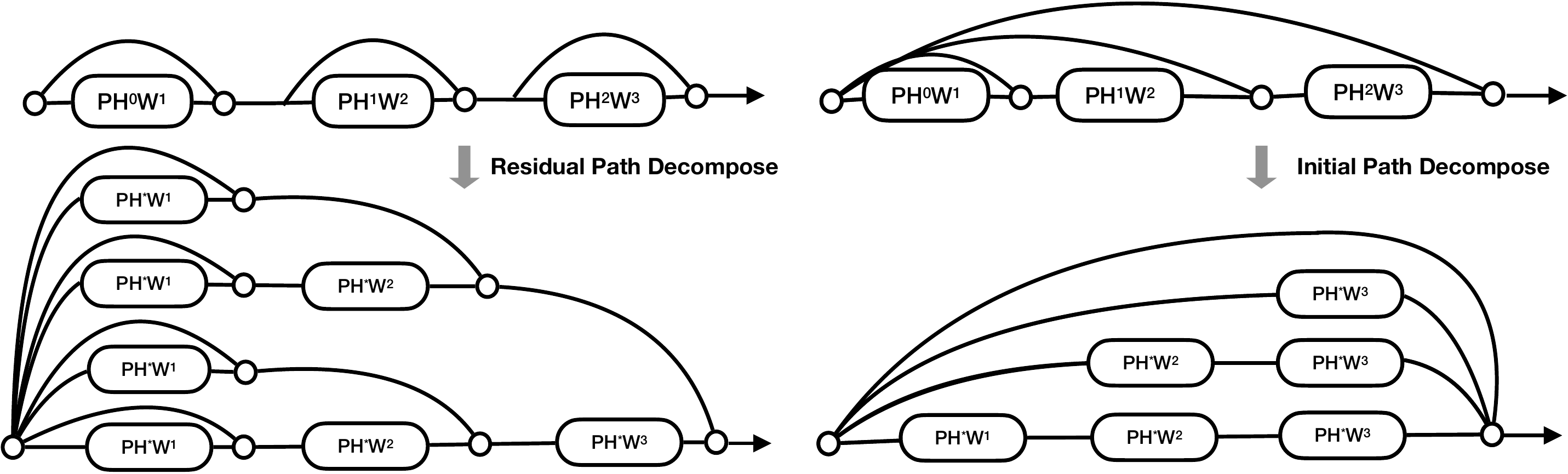}}
\caption{The path decomposition of GNNs with residual and initial connections.}
\label{fig:path}
\end{figure}
\subsection{The Unified Decomposition View of GNNs with Skip Connections}
Inspired by the ensemble view of ResNet~\cite{veit2016residual}, we decompose the path for GNNs from an ensemble view. We start by GNNs with residual and initial connections with three building blocks from inputs $\mathbf{H^0}$ to $\mathbf{H^3}$ and illustrate the decomposition tree of the path in Figure~\ref{fig:path}. For simplicity, we ignore the non-linear activation function ReLU since it does not influence the performance of GNNs as reported in ~\cite{chen2020simple,wu2019simplifying,zhou2020towards}.
The $L$ layers GNNs with skip connections can be seen as ensembles paths of different lengths (from 0 to $L$), and the length of each individual path is equivalent to the propagation times of graph convolutions. Therefore, we can formalize the general path decomposition of GNNs with skip connections as follows.

\begin{theorem}\label{theorem:foward_path_decomposition}(Path decomposition of GNNs with skip connection) The output of a depth $L$ GNNs with skip connection and input $H^0$ is given by
\begin{align}
\mathbf{H}^L &= \textstyle{\sum_{path \in Path}}\mathbf{P}_{path}\mathbf{H}^0\mathbf{W}_{path},
\end{align}
where $\textstyle{\mathbf{P}_{path}=\mathbf{P}^L_{path}...\mathbf{P}^1_{path}}$ is the product of the propagation matrix $P$ in GNNs, and $\textstyle{\mathbf{W}_{path}=\mathbf{W}^1_{path}...\mathbf{W}^L_{path}}$ is the weight matrix. We have $\textstyle{\mathbf{P}^l_{path}=\mathbf{P}}$ and $\textstyle{\mathbf{W}^l_{path}=\mathbf{W}^l}$ if unit $l$ selects the graph convolution operator in the current path; otherwise, they would degenerate to an identity matrix for the shortcut connection.
\end{theorem}

Note that in the ensemble view of ResNet, most paths are assumed to be positive for the final prediction since they represent the multi-scale resolution feature for one sample (images). 
However, due to the stacking of the propagation matrix P, the longer path in GNNs may contain an exponential number of neighbors' information and cause over-smoothing, which may hurt the discrimination ability of samples (nodes on the graph). 
\subsection{Forward Inference Path Decomposition}\label{sec:forward}
\textbf{Forward Path of standard GNNs.} We consider the path of SGC without the weight matrix $\mathbf{W}$ and the GCN with $\mathbf{W}$. Obviously, the SGC and GCN only have a single path, and due to the stacking of propagation matrix P, the node representation tends to be over-smoothed when the length of the path increases.
\begin{align}
    \text{(SGC): } H^L = PH^{L-1} = P^{L}H^0, \quad
    \text{(GCN): } H^L = PH^{L-1}W^l = P^{L}H^0{\textstyle\prod_{l=0}^L}W^l.
\end{align}
According to previous studies~\cite{liu2020towards,oono2019graph}. For the SGC, the stacking exponential of propagation matrix $\mathbf{P}$ is similar to the Markov process and may inevitably result in the node embedding converging to the degree eigenvector, which causes the node to be indistinguishable. The GCN, which stacks the transformation matrix $\mathbf{W}$ on the right side, also exponentially converges to over-smoothing according to the subspace theorem. 

\textbf{Forward path of GNNs with residual connections.} When considering the forward path decomposition for the residual connections, each unit can choose to propagate neighbor information or pass through the short cut. Therefore, the length of the path follows the binomial distribution, and the medium-length paths contribute noticeably, i.e., these paths would become over-smoothed when increasing graph convolution layers and dominate the last layer representation.
\begin{align}
     \text{(SGC w/residual): } H^L &= PH^{L-1} + H^{L-1} = (P+I)^LH^{0}=\textstyle{\sum_{l=0}^L} \textstyle{\binom{L}{l}} P^{l}H^0,\\
     \text{(GCN w/residual): } H^L &= PH^{L-1}W^l + H^{L-1} =  \textstyle{\sum_{path \in Path}} {P}_{path}H^0 \textstyle{\prod_{j\in path}^{}}{W}^j.
\end{align}
Recently, as reported in~\cite{chen2020simple,xu2018representation}, the behavior of $(P+I)^L$ in SGC is equivalent to a lazy random walk, which eventually converges to the stationary state and thus leads to over-smoothing. However, they omit the influence of the weight matrix $\mathbf{W}$.  On the other hand, the convergence proof of the GCN with residual connection and $\mathbf{W}$ was provided in \cite{oono2019graph}. However, since the upper bound of the convergence rate is larger than 1, there is no guarantee that expressive power will be lost~\cite{cong2021provable}.

Recall the motivation behind ResNet is the ability of helping the network in optimizing $\mathbf{W}$ for identity mapping. For the GCN with weight matrix $W$, consider the following scenario: if all the weight matrix $W$ can learn to be zero, the whole routes degenerate to a single pass-through from inputs to output, which can achieve the identity mapping and thus prevent over-smoothing induced by the graph convolution.
As shown in Figure~\ref{fig:oversmooth}(b), compared with the SGC, the residual connection significantly improves the performance of GCN on Cora-Random, which means that the optimization of weight matrix $W$ can indeed achieve identity mapping in shallow layers.
However, performance degradation of GNNs with residuals occurs even with the normalization technique~\cite{chen2020simple,li2020deepergcn}, as the number of layers rises. 
Therefore, we postulate that the character of the DeepGNNs, namely the successive entanglement of $P$ and $W$, causes the optimization difficulty for identity mapping. We will further discuss this point in Section~\ref{sec:backward}.

\textbf{Forward path of GNNs with initial connection.} The initial connection explicitly combines the initial node embedding with each layer to preserve the distinguishability of nodes. Moreover, unlike residual connections, the length of paths follows a uniform distribution. Therefore, the smooth representations of the longer paths cannot easily dominate the final output when increasing the number of layers.
\begin{align}
    \text{(SGC w/initial): } H^L &=PH^{L-1}+H^0= \textstyle{\sum_{l=0}^L} P^{l}H^0  \\
    \text{(GCN w/initial): } H^L &= PH^{L-1}W^L+H^0 = \textstyle{\sum_{l=0}^L} P^{l}H^0\textstyle{\prod_{i=L-l}^L} W^i
\end{align}
However, as shown in Figure~\ref{fig:path}, the paths of initial connections are the subset of residual connections, which implies that the residual connection may has more expressive capacity. Moreover, in the domain of other deep architectures~\cite{he2016deep,he2016identity,vaswani2017attention}, the residual connection is also more widely used than the initial connection. 
In the following, to unlock the potential of residual connections, which is limited by the forward over-smoothing, we will investigate the backward gradient for the optimization of residuals in GNNs.

\subsection{Gradient Smoothness Analysis from Backward Path Decomposition}\label{sec:backward}
Here, we give another analysis from backward path decomposition to investigate the reason why the residual connections fail in GNNs. Since the optimization of parameter $\mathbf{W}$ in ResNet is the key to learning the identity mapping, we take the initial step of analyzing the gradients in terms of parameters $\mathbf{W}$ and $\mathbf{H}$.

\begin{theorem}\label{theorem:backward_path_decomposition}(Backward gradient path decomposition of GNNs with skip connection). Given the training cross-entropy loss $\mathcal{L}$, the gradient in graph convolution with respect to parameter $\mathbf{W}^l$ is:

\begin{align}
\frac{\partial \mathcal{L}}{\partial \boldsymbol{W}^{l}}=&\left(\boldsymbol{P}\boldsymbol{H}^{l-1}\right)^{\top} \cdot \textstyle{\sum_{path\in Path}^{l\longrightarrow L}} {\boldsymbol{P}}_{path}^{\top} \cdot \frac{\partial \mathcal{L}}{\partial \boldsymbol{H}^{L}} \cdot \boldsymbol{W}_{path}^{\top}, 
\end{align}
where $\textstyle{\frac{\partial \mathcal{L}}{\partial \boldsymbol{H}^{L}}}$ indicates the initial gradient of the last layer. The details and proof of each GNN with skip connections can be found in the Appendix.
\end{theorem}

Note that the backward gradients are backpropagated through the neighborhood aggregation ($\mathbf{P}$) and feature transformation ($\mathbf{W}$), which are similar to the forward inference process. Specifically, the initial gradient signal would be smoothed due to the smoothness representation $\mathbf{H}^L$. Then, the product of ${\boldsymbol{P}}$s on the paths further smooths the gradient and causes the structural information loss of the gradient. Therefore, it was difficult for GNNs with residual connections to be optimized appropriately.
Next, we show the gradient path of the GCN, ResGCN and InitGCN. 

\textbf{Gradient Path of GCN (with/without) initial connection.} The backward gradient of GCN layers with or without an initial connection is the same, which implies that the W in the shallow layer of the initial connection is also difficult to optimize. 
\begin{align}
\label{eqn:gradient_of_initial}
\frac{\partial \mathcal{L}}{\partial \boldsymbol{W}^{l}}=&\left(\boldsymbol{H}^{l-1}\right)^{\top}\left({\boldsymbol{P}}^{L-l+1}\right)^{\top} \frac{\partial \mathcal{L}}{\partial \boldsymbol{H}^{L}} \cdot(\textstyle{\prod_{j=l+1}^L}\boldsymbol{W}^{j})^{\top}.
\end{align}

\textbf{Gradient Path of GCN with residual connection.} The backward gradient of the GCN with residual connections is analogous to the forward inference. The gradient of shallow layers is a collection of longer paths, which would also be smoothed.
\begin{align}
\frac{\partial \mathcal{L}}{\partial \boldsymbol{W}^{l}}=&\left(P\boldsymbol{H}^{l-1}\right)^{\top} \cdot \textstyle{\sum_{path\in Path}^{l\longrightarrow L}} \left({\boldsymbol{P}}^{path}\right)^{\top} \frac{\partial \mathcal{L}}{\partial \boldsymbol{H}^{L}}(\prod_{j\in path}\boldsymbol{W}^j)^{\top}. 
\end{align}

\begin{figure}[t]
\centering
\centerline{\includegraphics[width=\linewidth]{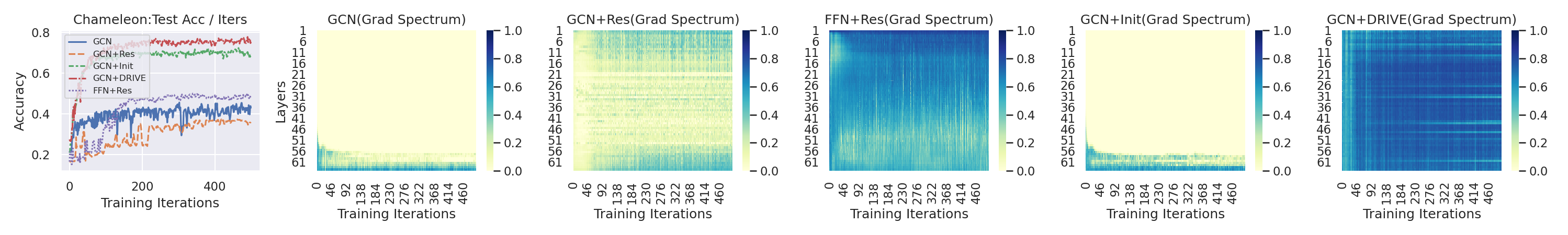}}
\caption{The grad spectral property (von Neumann entropy) of different skip connection variants of 64 graph convolutional layers during training over the Chameleon dataset. More details can be found in the Appendix.}
\label{fig:grad}
\end{figure}
To show the evolution of the gradient smoothness during training, we utilize the von Neumann entropy~\cite{bengtsson2017geometry} to calculate the spectral information of the gradient matrix and visualize it in Figure~\ref{fig:grad}. We can see that the evolution of GCN and GCN+Init is similar, i.e., the structural information is lost at shallow layers, which verifies the backward gradient in Eq~(\ref{eqn:gradient_of_initial}). However, the superhighway from the initial node embedding to deeper layers maintains the performance of the initial connection. For the residual connection, we first notice that, compared with FFN+Res, the gradient of GCN+Res also loses structural information. This causes difficulty in the optimization of GCN+Res. Therefore, the gradient smoothness problem and optimization difficulty for $\mathbf{W}$ are the keys to the performance degeneration for GNNs with skip connections.
\section{Universal Deep GNNs Framework}
\subsection{DRIVE Connections for Anti-Oversmoothing}
According to our analysis in Section~\ref{sec:forward}, the optimization difficulty is the key challenge for GNNs with residual connections to learn identity mapping and prevent over-smoothing. On the one hand, from the forward perspective, at the initial stage of training, the output $H$ of GNNs with the residual connection would be dominated by the over-smoothed longer path and cause the final representation to be indistinguishable, which also causes the initial gradient smoothing. On the other hand, from the backward perspective, the gradient becomes smoother due to the collection of stacking propagation matrix paths.
\begin{align}
\label{eqn:drive}
\mathbf{H}^{l+1} &= \mathbf{H}^{l}+ \alpha_l *\operatorname{GraphConv}(\mathbf{A}, \mathbf{H}^{l}) 
\end{align}

To overcome the over-smoothing of the output and gradient, we propose \textit{colD staRt adaptIVe rEsidual} (DRIVE) connections that initialize all graph convolution layers as the identity mappings, using additional learned parameters $\alpha_l$ for each layer with {\it zero} initialization. The {\it cold start} initialization deactivates all neighbor information propagation paths at the beginning of GNNs to prevent output and gradient smoothing. Moreover, it adaptively controls the smoothness weight to fit the downstream task during training, e.g., to deal with the homophily or heterophily graph. This behavior is similar to freezing the GraphConv at the initial stage and then warming it up to control smoothness during the following training, thus, termed {\it cold-start adaptive residual}. In the experimental part, we also visualize how the magnitude of the parameter $\alpha$s evolve in the training process.

\subsection{UDGNN with FeedForward Module}
Generally, in traditional deep learning, stacking multiple feature transformations can better learn the representative features and fit the data distribution~\cite{lu2020universal,qi2017pointnet}. 
However, current GNNs always entangle the propagation with transformation, which may only achieve the goal of the token mixer to some extent.

\begin{wrapfigure}{r}{0.2\textwidth}
\vspace{-0.5cm}
  \begin{center}
    \includegraphics[scale=0.7]{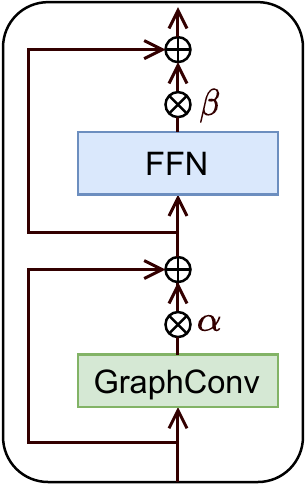}
  \end{center}
  \caption{Building block of UDGNN with DRIVE connections.}
\vspace{-1.2cm}
 \label{fig:udgnn}
\end{wrapfigure}

Inspired by the network architectures design of transformers~\cite{vaswani2017attention,yu2021metaformer}, we further introduce the channel FeedForward Module (FFN) with the proposed DRIVE connection after each graph convolution layer to increase the fitting power of GNNs. We combine these two into a Universal Deep GNNs framework (UDGNN) and use the UDGNN* to indicate the one without the FFN as Eq~(\ref{eqn:drive}). As shown in Figure~\ref{fig:udgnn}, where the encoder and decoder can be implemented with a linear layer, and the UDGNN is a conv-agnostic framework and can easily enhance the performance of GNNs in various datasets by stacking.
\begin{align}
\mathbf{H}^0 &= \operatorname{Encoder}(\mathbf{X}), \mathbf{Y} = \operatorname{Decoder}(\mathbf{H}^{L})  \\
\mathbf{M}^{l} &= \mathbf{H}^{l}+ \alpha_l *\operatorname{GraphConv}(\mathbf{A}, \mathbf{H}^{l}) \\ 
\mathbf{H}^{l+1} &= \mathbf{M}^l+ \beta_l *\operatorname{FFN}(\mathbf{M}^{l}) 
\end{align}

\section{Experiments}
\label{sec:experiments}
In this section, we aim to answer the following questions. RQ1: Can UDGNNs improve the performance of the standard GNNs on different kind of datasets?
RQ2: Compared with other techniques, does the DRIVE connection help train DeepGNNs?
RQ3: How does the behavior of the residual weight $\alpha$ and $\beta$. More experimental results can be found in the Appendix.

\subsection{Datasets and experimental settings}
\noindent
\textbf{Datasets.} 
We evaluate the performance on eleven well-known real-world datasets, which are summarized in the Appendix, including four homophily datasets and seven heterophily datasets. For all benchmarks, we use the feature vectors, class labels, and standard data splits from~\cite{hu2020open,lim2021large,pei2019geom,zhu2020beyond}.
\noindent
 \textbf{Baselines.} We compare our method with the following baselines: \textbf{(1)} Classical GNNs: GCN~\cite{GCN}, GAT~\cite{GAT} and GraphSAGE~\cite{Hamilton2017InductiveRL}; The basic MLP and the first heterophily GNNs GEOM-GCN~\cite{pei2019geom};
 \textbf{(2)} recent state-of-the-art decoupling GNNs for tackling heterophily and oversmooth:
 MixHop~\cite{abu2019mixhop}, GPRGNN~\cite{chien2020adaptive},
 APPNP~\cite{chien2020adaptive},
 DAGNN~\cite{chien2020adaptive},
 \textbf{(3)} recent state-of-the-art DeepGNNs with skip connections:
  FAGCN~\cite{bo2021beyond}, GCNII~\cite{chen2020simple};
  DeeperGNNs~\cite{li2021deepgcns},
  JKNet~\cite{xu2018representation},
  H2GCN~\cite{zhu2020beyond},
 For ease of comparison, we copy the reported results in literature~\cite{hu2020open,lim2021large,pei2019geom,zhu2020beyond}. Moreover, for the missing results under these splits, we rerun their released code with the best hyper-parameter over 10 times.

\noindent
\textbf{Parameters Setting.}
We adopt the same settings of the default hyper-parameters for UDGNNs and corresponding baselines (GCN, GAT and GraphSAGE), i.e., 128 hidden dimensions for OGB datasets and 64 hidden dimensions for others. We report the best best performance of UDGNNs between 2-64 layers for each dataset.  
For other baselines, we use their best default parameters and layers reported in the original papers. The detailed information is reported in the Appendix.

\subsection{Comparison with SOTA on Homophily \& Heterophily Datasets (RQ1)}
\setlength{\tabcolsep}{3pt}
\begin{table*}[t]
   \scriptsize
    \centering  
    \caption{Mean test accuracy $\pm$ stdev on 6 heterophily and 3 homophily real-world datasets over 10 fixed splits (48\%/32\%/20\% of nodes per class for train/val/test). The best performance is highlighted. $\ddagger$ denotes the results obtained from~\protect\cite{zhu2020beyond}.}
    \label{tab:ssnc-results}
    \begin{adjustbox}{width=\textwidth}
    \begin{tabular}{l|cccccc|ccc|c} %
    \toprule
       & \texttt{\bf Texas}& \texttt{\bf Wisconsin}& \texttt{\bf Actor} & \texttt{\bf Squirrel} & \texttt{\bf Chameleon} &
       \texttt{\bf Cornell}& \texttt{\bf Citeseer} &   \texttt{\bf Pubmed} &   \texttt{\bf Cora} & {\textbf{Average}}\\
\midrule
    {GEOM-GCN$\ddagger$} & $67.57$ & $64.12$ & $31.63$ & $38.14$ & $60.90$ & $60.81$ 
	   & \cellcolor{blue!15}${77.99}$ &  \cellcolor{blue!15}${90.05}$ & $85.27$ & 64.05 \\
	   {MLP$\ddagger$} & $81.89{\scriptstyle\pm4.78}$ & $85.29{\scriptstyle\pm3.61}$ & $35.76{\scriptstyle\pm0.98}$ & $29.68{\scriptstyle\pm1.81}$ & $46.36{\scriptstyle\pm2.52}$ & $81.08{\scriptstyle\pm6.37}$ & 
	   $72.41{\scriptstyle\pm2.18}$ & $86.65{\scriptstyle\pm0.35}$ & $74.75{\scriptstyle\pm2.22}$ &  65.99\\
	   \midrule
	   {MixHop$\ddagger$} & $77.84{\scriptstyle\pm7.73}$ & $75.88{\scriptstyle\pm4.90}$ & $32.22{\scriptstyle\pm2.34}$ & $43.80{\scriptstyle\pm1.48}$ & $60.50{\scriptstyle\pm2.53}$ & $73.51{\scriptstyle\pm6.34}$ & 
	   $76.26{\scriptstyle\pm1.33}$ & $85.31{\scriptstyle\pm0.61}$ & ${87.61}{\scriptstyle\pm0.85}$ & 68.21\\
	   {GPRGNN} & $82.12{\scriptstyle\pm7.72}$ & $81.16{\scriptstyle\pm3.17}$ & 
	   $33.29{\scriptstyle\pm1.39}$ & 
	   $43.29{\scriptstyle\pm1.66}$ & 
	   $61.82{\scriptstyle\pm2.39}$ &  
	   $81.08{\scriptstyle\pm6.59}$ &  
	   $75.56{\scriptstyle\pm1.62}$ & 
	   $86.85{\scriptstyle\pm0.46}$ & 
	   $86.98{\scriptstyle\pm1.33}$ & 70.15
	   \\ 
       {APPNP} & $78.37{\scriptstyle\pm6.01}$ & $81.42{\scriptstyle\pm4.34}$ & $34.64{\scriptstyle\pm1.51}$ & $33.51{\scriptstyle\pm2.02}$ & $47.50{\scriptstyle\pm1.76}$ & $77.02{\scriptstyle\pm7.01}$ & 
	   $77.06{\scriptstyle\pm1.73}$ & $87.94{\scriptstyle\pm0.56}$ & $87.71{\scriptstyle\pm1.34}$ &  67.24\\
	   {DAGNN} & $70.27{\scriptstyle\pm4.93}$ & $71.76{\scriptstyle\pm5.25}$ &$35.51{\scriptstyle\pm1.10}$ &$30.29{\scriptstyle\pm2.23}$ & $45.92{\scriptstyle\pm2.30}$ &	$73.51{\scriptstyle\pm7.18}$  &$76.44{\scriptstyle\pm1.97}$ 	&$89.37{\scriptstyle\pm0.52}$ 	 & $86.82{\scriptstyle\pm1.67}$ & 64.43\\
       \midrule
       {H2GCN-1$\ddagger$} & 
       $84.86{\scriptstyle\pm6.77}$ & 
       ${86.67}{\scriptstyle\pm4.69}$ & $35.86{\scriptstyle\pm1.03}$ & $36.42{\scriptstyle\pm1.89}$ & $57.11{\scriptstyle\pm1.58}$ & $82.16{\scriptstyle\pm4.80}$ & 
       $77.07{\scriptstyle\pm1.64}$ & $89.40{\scriptstyle\pm0.34}$ & $86.92{\scriptstyle\pm1.37}$ & 70.72\\
      {H2GCN-2$\ddagger$} & 
      $82.16{\scriptstyle\pm5.28}$ & $85.88{\scriptstyle\pm4.22}$ & $35.62{\scriptstyle\pm1.30}$ & $37.90{\scriptstyle\pm2.02}$ & $59.39{\scriptstyle\pm1.98}$ & $82.16{\scriptstyle\pm6.00}$ & 
      $76.88{\scriptstyle\pm1.77}$ & $89.59{\scriptstyle\pm0.33}$ & $87.81{\scriptstyle\pm1.35}$ & 70.87\\

	   {JKNet-GCN$\ddagger$} & $66.49{\scriptstyle\pm6.64}$ & $74.31{\scriptstyle\pm6.43}$ & $34.18{\scriptstyle\pm0.85}$ & $40.45{\scriptstyle\pm1.61}$ & $63.42{\scriptstyle\pm2.00}$ & $64.59{\scriptstyle\pm8.68}$ & 
	   $74.51{\scriptstyle\pm1.75}$ & $88.41{\scriptstyle\pm0.45}$ & $86.79{\scriptstyle\pm0.92}$ & 65.79\\
	   {DeeperGCN} &$70.27{\scriptstyle\pm7.09}$	&$72.75{\scriptstyle\pm4.84}$	&$35.57{\scriptstyle\pm1.08}$	&$31.23{\scriptstyle\pm1.35}$	&$48.75{\scriptstyle\pm2.57}$	&$68.38{\scriptstyle\pm5.85}$	&$75.58{\scriptstyle\pm1.18}$	&$88.80{\scriptstyle\pm0.40}$	&$85.61{\scriptstyle\pm1.94}$ & 64.10\\
	   {FAGCN} & $78.11 {\scriptstyle\pm 5.01}$ & $81.56{\scriptstyle\pm4.64}$ & $35.41{\scriptstyle\pm1.18}$ & $42.43{\scriptstyle\pm2.11}$ & $56.31{\scriptstyle\pm3.22}$ & $76.12{\scriptstyle\pm7.65}$ & 
	   $74.86{\scriptstyle\pm2.42}$ & 
	   $85.74{\scriptstyle\pm0.36}$ & $83.21{\scriptstyle\pm 2.04}$ & 68.18  \\
	   {GCNII} & $69.72{\scriptstyle\pm8.90}$ & $75.29{\scriptstyle\pm4.64}$ & $35.58{\scriptstyle\pm1.25}$ & $47.21{\scriptstyle\pm1.73}$ & $60.79{\scriptstyle\pm2.35}$ & $79.19{\scriptstyle\pm6.12}$ & $76.82{\scriptstyle\pm1.67}$ & $89.26{\scriptstyle\pm0.48}$ & \cellcolor{blue!15}$87.89{\scriptstyle\pm1.88}$ & 69.07 \\

       \midrule
       {GraphSAGE$\ddagger$} & $82.43{\scriptstyle\pm6.14}$ & $81.18{\scriptstyle\pm5.56}$ & $34.23{\scriptstyle\pm0.99}$ & $41.61{\scriptstyle\pm0.74}$ & $58.73{\scriptstyle\pm1.68}$ & $75.95{\scriptstyle\pm5.01}$ & 
	   $76.04{\scriptstyle\pm1.30}$ & $88.45{\scriptstyle\pm0.50}$ & $86.90{\scriptstyle\pm1.04}$ & 69.50 \\
	   \textbf{$\text{UDGNN*}_{SAGE}$} &  $82.97{\scriptstyle\pm3.87}$ &  $85.55{\scriptstyle\pm4.89}$ & $36.38{\scriptstyle\pm1.52}$ & $62.09{\scriptstyle\pm2.16}$ & $71.05{\scriptstyle\pm1.83}$ & $82.16{\scriptstyle\pm6.88}$ & $76.35{\scriptstyle\pm1.69}$ & $89.60{\scriptstyle\pm0.36}$ & $86.71{\scriptstyle\pm1.18}$ & 74.76\\
	   \textbf{$\text{UDGNN}_{SAGE}$} &  $84.05{\scriptstyle\pm4.11}$ &  $86.86{\scriptstyle\pm4.42}$ & \cellcolor{blue!15}$36.64{\scriptstyle\pm1.18}$ & $62.02{\scriptstyle\pm2.03}$ & $69.51{\scriptstyle\pm2.22}$ & $83.24{\scriptstyle\pm7.83}$ & $75.85{\scriptstyle\pm1.69}$ & $89.88{\scriptstyle\pm0.41}$ & $86.65{\scriptstyle\pm1.18}$ & 74.97\\
       \midrule
	   {GAT$\ddagger$} & $58.38{\scriptstyle\pm4.45}$ & $55.29{\scriptstyle\pm8.71}$ & $26.28{\scriptstyle\pm1.73}$ & $30.62{\scriptstyle\pm2.11}$ & $54.69{\scriptstyle\pm1.95}$ & $58.92{\scriptstyle\pm3.32}$ & 
	   $75.46{\scriptstyle\pm1.72}$ & $84.68{\scriptstyle\pm0.44}$ & $82.68{\scriptstyle\pm1.80}$ & 58.56 \\
	   \textbf{$\text{UDGNN*}_{GAT}$} & $80.27{\scriptstyle\pm4.23}$ & $83.72{\scriptstyle\pm4.43}$ & $35.72{\scriptstyle\pm1.57}$ & \cellcolor{blue!15} $66.21{\scriptstyle\pm1.79}$ & $71.36{\scriptstyle\pm1.68}$ & $81.89{\scriptstyle\pm6.11}$ & $74.92{\scriptstyle\pm1.47}$ & $89.62{\scriptstyle\pm0.47}$ & $85.55{\scriptstyle\pm1.41}$ & 74.36\\
	   \textbf{$\text{UDGNN}_{GAT}$} &
    $83.43{\scriptstyle\pm4.33}$ & $85.47{\scriptstyle\pm3.97}$ & 
    $36.13{\scriptstyle\pm1.02}$ & 
     $63.41{\scriptstyle\pm1.84}$ & 
     $68.15{\scriptstyle\pm1.54}$ & 
     $82.71{\scriptstyle\pm4.06}$ & $75.09{\scriptstyle\pm1.75}$ & $89.78{\scriptstyle\pm0.56}$ & $85.37{\scriptstyle\pm1.25}$ & 74.39\\
       \midrule
	   {GCN$\ddagger$} & $59.46{\scriptstyle\pm5.25}$ & $59.80{\scriptstyle\pm6.99}$ & $30.26{\scriptstyle\pm0.79}$ & $36.89{\scriptstyle\pm1.34}$ & $59.82{\scriptstyle\pm2.58}$ & $57.03{\scriptstyle\pm4.67}$ & 
	   $76.68{\scriptstyle\pm1.64}$ & $87.38{\scriptstyle\pm0.66}$ & $87.28{\scriptstyle\pm1.26}$ &  61.62\\
    \textbf{$\text{UDGNN*}_{GCN}$} & $81.25{\scriptstyle\pm6.78}$ & $86.47{\scriptstyle\pm4.34}$ & $35.43{\scriptstyle\pm0.96}$ & \cellcolor{blue!15}$70.05{\scriptstyle\pm2.24}$ & 
    \cellcolor{blue!15}$76.79{\scriptstyle\pm1.46}$ & $82.43{\scriptstyle\pm5.73}$ & $76.15{\scriptstyle\pm1.64}$ & $89.39{\scriptstyle\pm0.45}$ & $87.28{\scriptstyle\pm0.89}$ &  76.14\\
    \textbf{$\text{UDGNN}_{GCN}$} & \cellcolor{blue!15}$84.60{\scriptstyle\pm5.32}$ & \cellcolor{blue!15}$87.64{\scriptstyle\pm3.74}$ & $36.13{\scriptstyle\pm1.21}$ & $68.13{\scriptstyle\pm2.59}$ & 
    $74.53{\scriptstyle\pm1.19}$ & \cellcolor{blue!15}$84.32{\scriptstyle\pm7.29}$ & $76.05{\scriptstyle\pm1.83}$ & $89.85{\scriptstyle\pm0.35}$ & $86.97{\scriptstyle\pm1.21}$ &  \cellcolor{blue!15}76.47\\
	   \bottomrule
    \end{tabular}
    \end{adjustbox}
\end{table*}

Table~\ref{tab:ssnc-results} and Table~\ref{tab:OGB} provide the accuracy of different GNNs on the node classification task over both homophily and heterophily datasets. We report the best performance of each model across different layers and summarize the following observation:
1) The performance of most GNNs on the homophily dataset is similar. They all outperform MLP models since the smooth process of graph convolution is helpful for these datasets with smooth signals. Moreover, our UDGNN can maintain the performance of the standard GNNs.
2) Due to the over-smoothing, most GNNs are worse than MLP when dealing with non-smooth heterophily datasets. Although decoupling designs can improve the performance of GNNs under heterophily to some extent, they remain shallow models without deep feature transformation and cannot achieve optimal results.
3) Compared with other SOTAs with skip connections and specific architectural modifications, our UDGNN* with just a simple drive connection can strongly improve the performance of the standard GNNs and achieve new state-of-the-art results. Notably, $\text{UDGNN*}_{GCN}$ achieves strong results on squirrel and chameleon datasets and outperforms previous state-of-the-art methods by a large margin, e.g., greater than 70\% accuracy. Moreover, the $\text{UDGNN}_{GCN}$ achieves the best average accuracy of 76.47\% on nine datasets and outperforms others on the OGB datasets, which demonstrate the effectiveness of our framework. 

The strong performance of the UDGNN reveals a surprising conclusion: we do not need specific architectural modifications for difficult heterophily datasets. Instead, just introducing a learnable parameter to the residual connection and allowing the GNNs to go deeper can well solve the over-smoothing and heterophily problem. 
Therefore, our proposed UDGNN could serve as a starting baseline for future GNN architecture designs.

\begin{wraptable}{r}{0.4\textwidth} 
{
\scriptsize
\vspace{-1.7cm}
\centering
\caption{Results on OGB Arxiv(Homophily) and Arxiv-year(Heterophily) over 10 runs.}\label{tab:OGB}
    \begin{tabular}{l|ll}
    \toprule
    Methods & Arxiv & Arxiv-year\\ \hline
    MLP &$55.50{\scriptstyle\pm0.23}$ & $36.70{\scriptstyle\pm0.21}$\\
    APPNP & $68.24{\scriptstyle\pm0.19}$ & $38.15{\scriptstyle\pm0.26}$\\ 
    GPR-GNN & $70.71{\scriptstyle\pm0.26}$ & $45.07{\scriptstyle\pm0.21}$\\  
    MixHop & $72.68{\scriptstyle\pm0.16}$ & $51.81{\scriptstyle\pm0.17}$\\  
    DAGNN & $72.09{\scriptstyle\pm0.25}$ & $42.76{\scriptstyle\pm0.26}$\\
    DeeperGCN & $71.92{\scriptstyle\pm0.16}$ & $43.45{\scriptstyle\pm0.25}$\\ 
    JKNet & $72.19{\scriptstyle\pm0.21}$ & $46.28{\scriptstyle\pm0.29}$ \\
    H2GCN & $72.28{\scriptstyle\pm0.17}$ & $49.09{\scriptstyle\pm0.10}$\\  
    GCNII & $72.74{\scriptstyle\pm0.16}$ & $47.21{\scriptstyle\pm0.28}$\\ 
    \midrule
    GCN & $71.74{\scriptstyle\pm0.29}$ & $46.02{\scriptstyle\pm0.26}$ \\ 
    $\text{UDGNN*}_{GCN}$ & ${72.82}{\scriptstyle\pm0.20}$ & 
    ${52.74}{\scriptstyle\pm0.23}$\\ 
    $\text{UDGNN}_{GCN}$ & 
    \cellcolor{blue!15}${72.94}{\scriptstyle\pm0.22}$ & 
    \cellcolor{blue!15}${53.16}{\scriptstyle\pm0.15}$\\ \hline
    GraphSAGE & $71.49{\scriptstyle\pm0.27}$ & $48.64{\scriptstyle\pm0.27}$\\  
    $\text{UDGNN*}_{SAGE}$ & ${72.23}{\scriptstyle\pm0.17}$&  ${51.86}{\scriptstyle\pm0.18}$\\ 
    $\text{UDGNN}_{SAGE}$ & ${72.34}{\scriptstyle\pm0.16}$ & ${52.38}{\scriptstyle\pm0.14}$\\ 
    \bottomrule
    \end{tabular}
    }
\vspace{-0.5cm}
\end{wraptable}


\subsection{Ablation Studies (RQ2)}
\begin{figure}[t]
\centering
\centerline{\includegraphics[width=\linewidth]{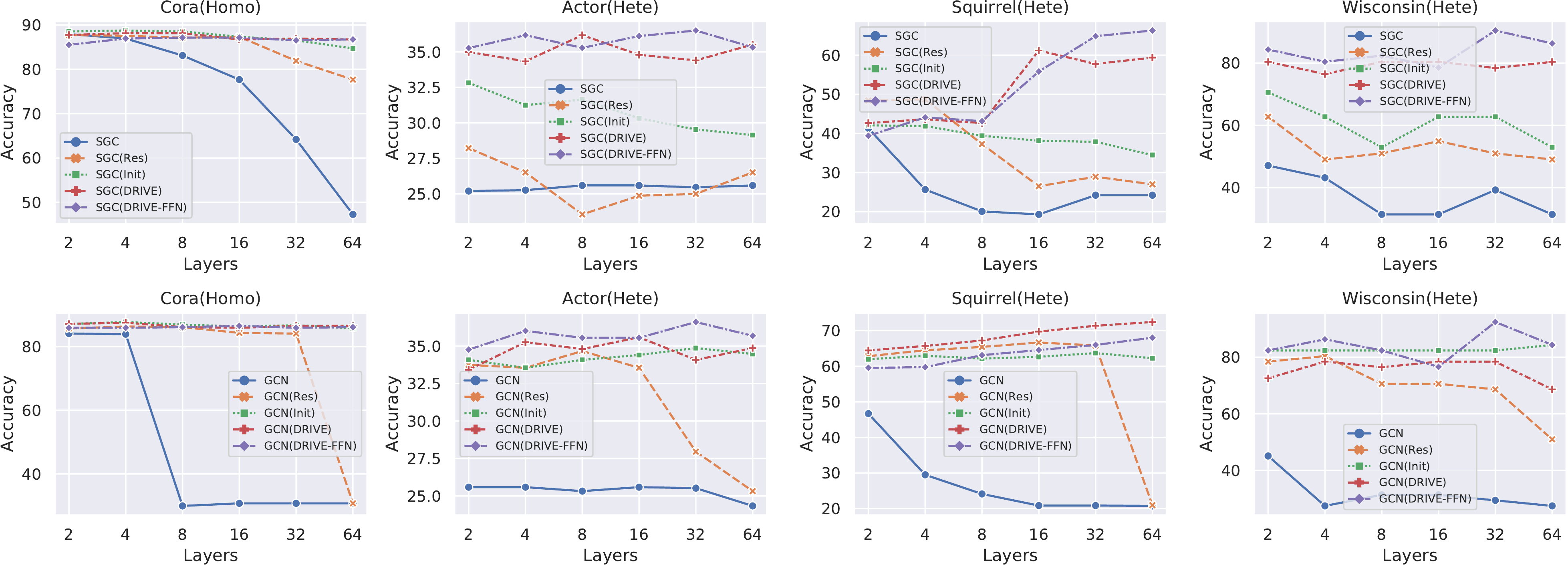}}
\caption{Ablation study on variants of skip connection on SGC and GCN under our framework in both homophily and heterophily datasets.
}
\label{fig:abalation}
\end{figure}

To gain deeper insight into the contributions of different variants of skip connections in our approach, we conduct experiments on varying UDGNN with 2 graph convolutions: SGC and GCN, and 5 skip connection variations: non-skip, residual, initial, DRIVE and DRIVE-FFN with additional FFN.
As shown in Figure~\ref{fig:abalation}, we make three observations: 
1) Both residual and initial connections can improve the performance on heterophily graphs at the shallow layer. Moreover, the GCN equipped with these connections outperforms the SGC counterparts since the weight matrices empower the GCN to learn identity mapping.
2) Compared with the residual connection, the initial connection significantly relieves over-smoothing. However, the results in the heterophily graphs are suboptimal, especially for the SGC.
3) The drive connection can successfully drive the deep graph residual to help the GCN and SGC effectively scale up the depth. Moreover, it achieves the best performance over heterophily graphs, and the FFN component slightly enhances performance on some datasets. Notice that we do not apply any normalization or regularization techniques designed for graphs.
This result suggests that the drive connection can solve the problem of over-smoothing and successfully train DeepGNNs to fit various smoothnesses of graph datasets.

\subsection{Visualization (RQ3)}
\begin{figure}[t]
\centering
\subfigure[Cora-Random (noisy neighbors)]
{
\begin{minipage}{0.45\linewidth}
\centering
\includegraphics[width=\linewidth]{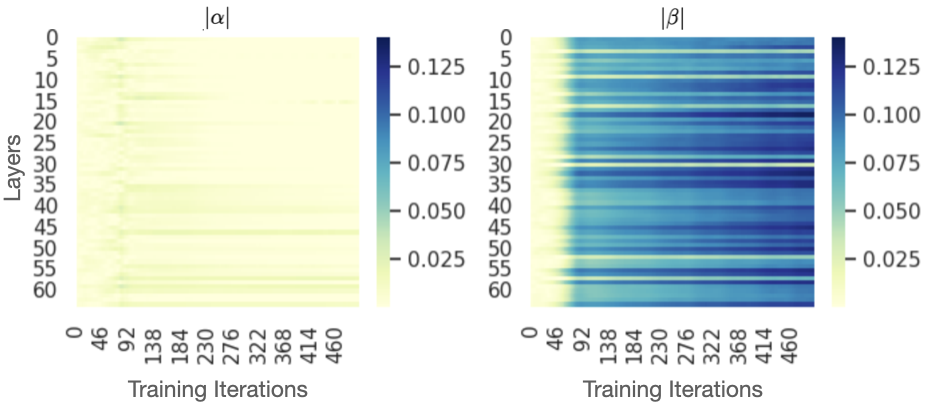}
\end{minipage}
}
\subfigure[Cora (helpful neighbors)]
{
\begin{minipage}{0.45\linewidth}
\centering
\includegraphics[width=\linewidth]{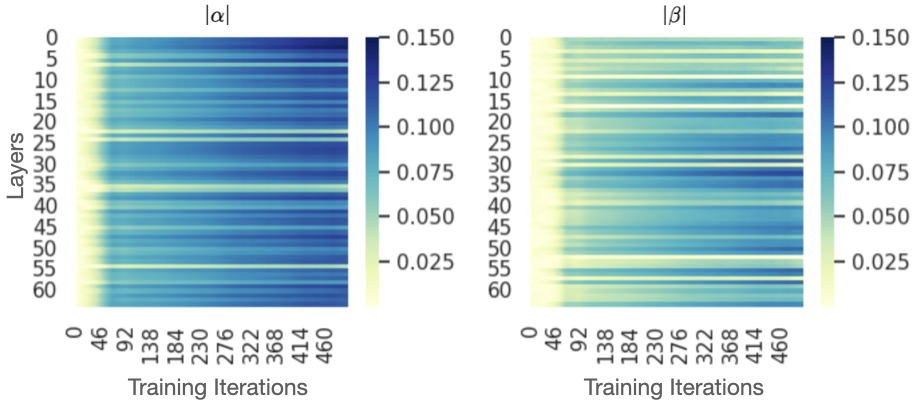}
\end{minipage}
}
\caption{The visualized evolution of $\left| \alpha_l \right|$ and $\left| \beta_l \right|$ in the training process for 64 layers UDGNN.}
\label{fig:visual_of_alphas}
\end{figure}

To further understand how the weights $ \alpha_l$ and $ \beta_l $ are learned in the training process. We visualize their dynamics of the 64 layers UDGCN using the Cora and Cora-Random (fully connected graph) datasets in Figure~\ref{fig:visual_of_alphas}. 
Notice that we use the {\it cold-start} to initialize $\alpha_l$ and $\beta_l$ to help the model learn the non-smooth node representation by preventing information aggregation from neighbors. Then, in the training process, it automatic learns whether or not to aggregate information from each node's neighbors. As shown in Figure~\ref{fig:visual_of_alphas}, $\left | \alpha_l \right|$ remains a small value to maintain the identity mapping and discard the over-smoothed neighbor information for Cora-Random. In contrast, $\left | \alpha_l \right|$ of Cora dataset grows quickly for most layers.  For the other parameter $\left| \beta_l \right|$, since it controls the feature transformation of nodes individually, it is frozen initially and then grows quickly regardless of the neighbor information. 

\section{Conclusion}\label{sec:conclusion}
In this work, we have examined the forward and backward behavior of GNNs with residual connections from a novel path decomposition perspective. We discovered that the gradient smoothing problem in the backward gradient optimization of GNNs with residual connections prevents the model from optimizing to learn the identity mapping. In addition, we presented a framework for Universal Deep GNNs that provides standard GNNs with 
DRIVE connections for GraphConv and FeedForward module to become deeper and more powerful. Extensive experiments validate the effectiveness of our framework on homophily and heterophily graph datasets. Our proposed UDGNN may serve as a starting baseline for the architecture design of future GNNs.

%
\printbibliography

\clearpage
\appendix
\clearpage
\makeappendixtitle

\noindent\textbf{Organization.}
In Section~\ref{supp:data_statistics}, we provide the details of the data statistics.
In Section~\ref{supp:implement_detail}, we provide the implementation details and hyper-parameters of UDGNNs.
In Section~\ref{supp:proof_of_theorem}, we provide the proofs of Theorem~\ref{theorem:foward_path_decomposition} and Theorem~\ref{theorem:backward_path_decomposition}, respectively.
In Section~\ref{supp:path_empirical_study}, we provide more empirical results for the path decomposition view to illustrate the optimization difficulty and gradient smoothing.
In Section~\ref{supp:more_experiments}, we provide additional experiments to validate the effectiveness of the proposed DRIVE connection and the UDGNN framework.
In Section~\ref{supp:limitation}, we provide a discussion of limitations and potential negative impacts.
The code will be available at: 
\begin{center}
\url{https://github.com/JC-202/UDGNNs}.
\end{center}

\section{Data Statistics}\label{supp:data_statistics}

\setlength{\tabcolsep}{3pt}

\begin{table}[h]
\caption{Statistics of datasets.}
\label{tab:dataset}
\begin{tabular}{lccccccc}
\toprule
{Dataset} &{\#Classes} &{\#Nodes} &{\#Edges} &{\#Graph Type} &{\#Context} &{\#Features} &{\#Train/Val/Test}\\
\midrule
Texas &5 & 183& 309 &Heterophily &Web pages &1,703 & 48\%/32\%/20\% \\
Wisconsin &5 & 251& 499&Heterophily &Web pages &1,703 & 48\%/32\%/20\%\\
Cornell &5 & 183 & 295&Heterophily &Web pages &1,703 & 48\%/32\%/20\%\\ 
Squirrel &5 &5,201 & 217,073&Heterophily &Wiki pages &2,089 & 48\%/32\%/20\%\\ 
Chameleon & 5 & 2277 & 36,101&Heterophily &Wiki pages &2,325 & 48\%/32\%/20\%\\ 
Actor &5 &7,600 & 33,544&Heterophily &Movies &931 & 48\%/32\%/20\%\\ 
Arxiv-year & 5& 169,343  & 1,166,243 &Heterophily &OGB &128 & 50\%25\%25\% \\
Cora-Random &7 &2,708 &7,333,264 &Heterophily &Citation &1,433  & 5\%/19\%/37\%\\
\midrule
CiteSeer &6 &3,327 &4,732 &Homophily &Citation &3,703 & 48\%/32\%/20\% \\
PubMed &3 &19,717 &44,338 &Homophily &Citation &500 & 48\%/32\%/20\% \\
Cora &7 &2,708 &5,429 &Homophily &Citation &1,433  & 48\%/32\%/20\%\\
Arxiv  & 40 & 169,343 &1,166,243 &Homophily &OGB &128 & 48\%/32\%/20\%\\ 
\bottomrule
\end{tabular}
\end{table}

\begin{itemize}
    \item {Homophily Datasets}
    \begin{itemize}
        \item \textit{Citeseer, Pubmed, Cora}~\cite{GCN}: For the basic citation datasets, nodes correspond to papers; edges correspond to citation links, and the sparse bag-of-words are the feature representation of each node. Finally, the label of each node represents the topic of the paper. We use the fully supervised data split in~\cite{pei2019geom,zhu2020beyond}
        \item \textit{Arxiv}~\cite{hu2020open}: The Arxiv dataset is a large scale citation network collected from all Computer Science ARXIV papers. Each node is an ARXIV paper, and edges are citation relations between papers. The features are 128-dimensional averaged word embeddings of each paper, and labels are subject areas of papers.
    \end{itemize}
    \item {Heterophily Datasets}
    \begin{itemize}
        \item \textit{Texas, Wisconsin, Cornell}~\cite{pei2019geom}: Cornell, Texas, and Wisconsin are the web page networks captured from the computer science departments of these universities in the WebKB dataset. In these networks, nodes and edges represent the web pages and hyperlinks. Similar to the Citations networks, words in the web page represent the node features in the bag-of-word form. The web pages are labeled into five categories: student, project, course, staff, and faculty.
        \item \textit{Squirrel, Chameleon}~\cite{pei2019geom}: Chameleon and Squirrel are web pages extracted from different topics in Wikipedia. Similar to WebKB, nodes and edges denote the web pages and hyperlinks among them, respectively, and informative nouns in the web pages are employed to construct the node features in the bag-of-word form. Webpages are labeled in terms of the average monthly traffic level.
        \item \textit{Actor}~\cite{pei2019geom}: The actor network contains the co-occurrences of actors in films, which are extracted from the heterogeneous information networks. It describes the complex relationships among films, directors, actors and writers. In this network, nodes and edges represent actors and their co-occurrences in films, respectively. The actor’s Wikipedia page is used to extract features and node labels.
        \item \textit{Arxiv-year}~\cite{lim2021large}: Modifying node labels of the Arxiv dataset to the year of paper, and the goal is to predict the year of paper publication that allows for evaluation of GNNs in large scale non-homophilous settings.
        \item
        \textit{Cora-Random}: We construct the extremely heterophily scenario by connecting all nodes on the Cora dataset. We use the semi-supervised limited labeled data split in~\cite{GCN} to enhance the supervision difficulty. In the Cora-Random dataset, since the neighbor of each node provides no helpful information for classification, the standard graph convolution may cause over-smoothing rapidly due to the fully-connected property.
    \end{itemize}
\end{itemize}
\section{Implementation Details}\label{supp:implement_detail}

\noindent \textbf{Running environment:} All the GNN models implemented in PyTorch, and tested on a machine with 24 Intel(R) Xeon(R) CPU E5-2650 v4 @ 2.20 GHz processors, GeForce GTX-2080 Ti 11 GB GPU, NVIDIA A100 40 GB GPU, and 128 GB memory size.

\noindent \textbf{Hyper-parameters:}
For UDGNNs, we do not add the self-loop of graphs and employ the Adam optimizer and select the learning rate $\in \left \{ 0.001, 0.01, 0.05\right \}$, weight decay $\in \left \{ 0.00005, 0.0005 \right \}$ and dropout rate $\in \left \{ 0, 0.5 \right \}$ based on the validation sets.
For other models, we utilize their best default parameters in the original papers. 
\section{Proof of Theorem}\label{supp:proof_of_theorem} 

\noindent \textbf{Proof of Theorem~\ref{theorem:foward_path_decomposition}}

\begin{theorem*}(Path decomposition of GNNs with skip connection) The output of a depth $L$ GNN with skip connection and input $H^0$ is given by:
\begin{align}
\mathbf{H}^L &= \textstyle{\sum_{path \in Path}}\mathbf{P}_{path}\mathbf{H}^0\mathbf{W}_{path},
\end{align}
where $\textstyle{\mathbf{P}_{path}=\mathbf{P}^L_{path}...\mathbf{P}^1_{path}}$ is the product of the propagation matrix $P$ in GNNs, and $\textstyle{\mathbf{W}_{path}=\mathbf{W}^1_{path}...\mathbf{W}^L_{path}}$ is the weight matrix. We have $\textstyle{\mathbf{P}^l_{path}=\mathbf{P}}$ and $\textstyle{\mathbf{W}^l_{path}=\mathbf{W}^l}$ if unit $l$ selects the graph convolution operator in the current path; otherwise, they would degenerate to an identity matrix for the shortcut connection.
\end{theorem*}

\textit{Proof.} For simplicity, we assume the input vector $H^0$ to be non-negative and remove the ReLU activation $\sigma$ operation as~\cite{chen2020simple, wu2019simplifying}, then consider a three-layer GraphConv with a single path following from the Multiplication law that
\begin{align}
H^3=P\sigma(P\sigma(PH^0W^0)W^1)W^2=(PPP)H^0(W^0W^1W^2)=P_{path}H^0W_{path}.
\end{align}
Then, the proof follows from the fact that the output of standard GNNs with skip connections is formed by the summation of all the individual paths.

\noindent \textbf{Proof of Theorem~\ref{theorem:backward_path_decomposition}}

\begin{theorem*}(Backward gradient path decomposition of GNNs with skip connection). Given the training cross-entropy loss $\mathcal{L}$, the gradient in graph convolution with respect to parameter $\mathbf{W}^l$ is:

\begin{align}
\frac{\partial \mathcal{L}}{\partial \boldsymbol{W}^{l}}=&\left(\boldsymbol{P}\boldsymbol{H}^{l-1}\right)^{\top} \cdot \textstyle{\sum_{path\in Path}^{l\longrightarrow L}} {\boldsymbol{P}}_{path}^{\top} \cdot \frac{\partial \mathcal{L}}{\partial \boldsymbol{H}^{L}} \cdot \boldsymbol{W}_{path}^{\top}, 
\end{align}
where $\textstyle{\frac{\partial \mathcal{L}}{\partial \boldsymbol{H}^{L}}}$ indicates the initial gradient of the last layer.
\end{theorem*}

\textit{Proof.} Consider that the derivative $\cal{L}$ with respect to $W^l$ in GNNs with skip connections is:
\begin{align}
\label{eqn:derivative_Wl}
\frac{\partial \mathcal{L}}{\partial {W}^{l}}=&\frac{\partial H^l}{\partial {W}^{l}}\frac{\partial \mathcal{L}}{\partial {H}^{l}} =(PH^{l-1})^{\top}\cdot \frac{\partial \mathcal{L}}{\partial {H}^{l}}.
\end{align}
According to the forward path decomposition, the output of a depth $L$ GNN with skip connection and input $H^l$ is given by:
\begin{align}
H^L=&\textstyle{\sum_{path\in Path}^{l\longrightarrow L}} {\boldsymbol{P}}_{path}\cdot H^l \cdot \boldsymbol{W}_{path}
\end{align}
Therefore, following the fact that the derivative of $\frac{\partial \cal{L}}{\partial {H}^{l}}$ is the summation of all the individual paths. we can calculate the $\frac{\partial \cal{L}}{\partial {H}^{l}}$ by the chain rule and stacking the propagation matrix $P$ and the weight matrix $W$ in the $P_{path}$ and $W_{path}$ as the following equation:
\begin{align}
\label{eqn:derivative_Hl}
\frac{\partial \mathcal{L}}{\partial {H}^{l}}=&\textstyle{\sum_{path\in Path}^{l\longrightarrow L}} {\boldsymbol{P}}_{path}^{\top}\cdot \frac{\partial \mathcal{L}}{\partial {H}^{L}} \cdot \boldsymbol{W}_{path}^{\top}
\end{align}

Combining with Equation~\ref{eqn:derivative_Wl} and Equation~\ref{eqn:derivative_Hl}  complete the proof.

\section{More Empirical Study of Path Decomposition}\label{supp:path_empirical_study}
In this section, we report additional empirical results to illustrate the correctness of our analysis based on path decomposition.

\subsection{Forward Inference: Does residual connection help GraphConv learn the identity mapping?}

\begin{figure}[h]
\centering
\centerline{\includegraphics[width=\linewidth]{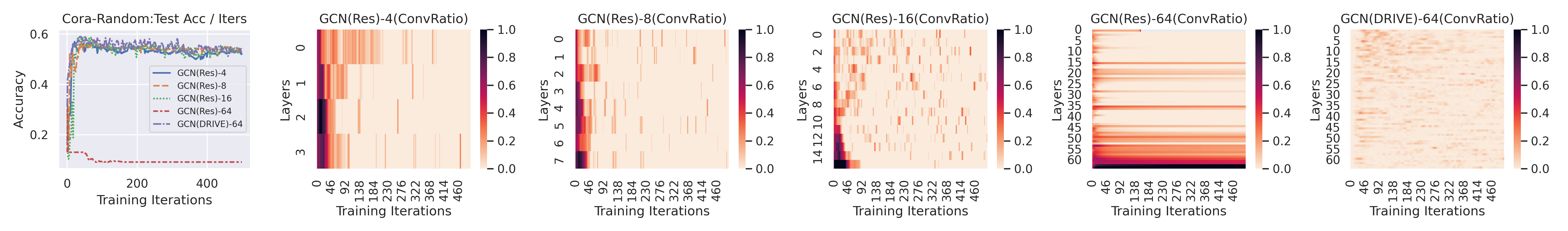}}
\caption{The test accuracy and ConvRatio for each layer during training over the Cora-Random dataset. We show the results of GCN(Res)-8, GCN(Res)-16, GCN(Res)-32, GCN(Res)-64, and GCN(DRIVE)-64. All variants of GraphConv with skip connections are under our encoder-decoder framework.}
\label{fig:forward_infer_identity}
\end{figure}

To investigate the ability to learn identity mapping, which is essential for the deep model to maintain performance as the shallow counterpart, we define the ConvRatio as the measurement of the difference between input and output for each layer. Compared with other vector similarity metrics, e.g., cosine similarity, the ConvRatio can capture the norm difference contributed by each graph convolution unit, which is important for identity mapping. The smaller of ConvRatio is, the better the identity mapping it learns.

\begin{align}
\text{ConvRatio}^l &= \sum_{i=0}^N\frac{\left \| h^{l+1}_i-h^{l}_i \right \|_2 }{N \cdot \left \| h^{l+1}_i \right \|}
\end{align}

Then, to show the ability to learn identity mapping of GraphConv with residual connection, we conduct experiments and visualize the evolution of ConvRatio over the Cora-Random dataset as in Figure~\ref{fig:forward_infer_identity}. We can see that the residual connection can indeed help the shallow GNNs model (4,8, and 16 layers) learn identity mapping, e.g., the ConvRatio tends to be 0 and test accuracy tends to 60 (the optimal result of the MLP), but it failed in the deeper GNNs (64 layers). This indicates that the residual connection can help the GNNs learn the identity mapping, preventing over-smoothing in the fully-connected Cora-Random dataset. However, it is difficult for the residual connection to learn identity mapping when GNNs become deeper. This motivates us to hypothesize that the optimization difficulty for deeper GCN(Res) prevents it from avoiding over-smoothing. In contrast, our proposed DRIVE connection works well in the deeper layers by solving the optimization difficulty. 

\subsection{Backward Gradient: Gradient Smoothness Phenomenon}

\subsubsection{von Neumann entropy for measuring the gradient matrix spectral information}
Compared with over-smoothing for the node features, backward propagation would result in losing structural information of gradient matrix when the gradient is over-smoothed. Hence, to measure the gradient smoothness, we consider the information of the gradient matrix from the spectral domain. Inspired by the von Neumann entropy in quantum statistical mechanics~\cite{bengtsson2017geometry}, which extends the idea of entropy for positive definite symmetric matrices to measure the matrices' information. Specifically, 
suppose $\sigma_1^l,\sigma_2^l, ...,\sigma_d^l $ denote singular values of weight matrix $W^l \in R^{d \times d}$, 
we then normalize them so that $\sum_{i=1}^{d}\sigma_i^l=1$, where $i=1,...,d$ for index of singular values.
The normalized von Neumann entropy of $W^l$ that represents gradient spectral information is computed by  
\begin{align}
{von(W^l)}=\frac{-\sum_{i=1}^{d}\sigma_i^l \log(\sigma_i^l) }{\log(d)}.
\end{align}

The above metric ranges from $[0,1]$ and can be used to quantify the spectral information and the smoothness of the gradient, i.e., a smaller of this metric indicates a smoother gradient.

\subsubsection{Evolution of gradient spectral information for different skip connections}

Then, to monitor the gradient smoothness during training, we visualize the evolution of the gradient spectral information in 64 layers of 5 variants under our framework, i.e., including GCN, GCN+Res, FFN+Res, GCN+Init and GCN+DRIVE. 
\begin{align}
H^{l+1} &= \sigma(PH^lW^l), &\text{1.GCN}\\
H^{l+1} &= H^l + \sigma(PH^lW^l), &\text{2.GCN+Res}\\
H^{l+1} &= H^l + \sigma(H^lW^l), &\text{3.FFN+Res}\\
H^{l+1} &= H^0 + \sigma(PH^lW^l), &\text{4.GCN+Init} \\
H^{l+1} &= H^l + \alpha_l*\sigma(PH^lW^l), &\text{5.GCN+DRIVE} 
\end{align}

\begin{figure}
     \centering
     \begin{subfigure}
         \centering
         \includegraphics[width=\textwidth]{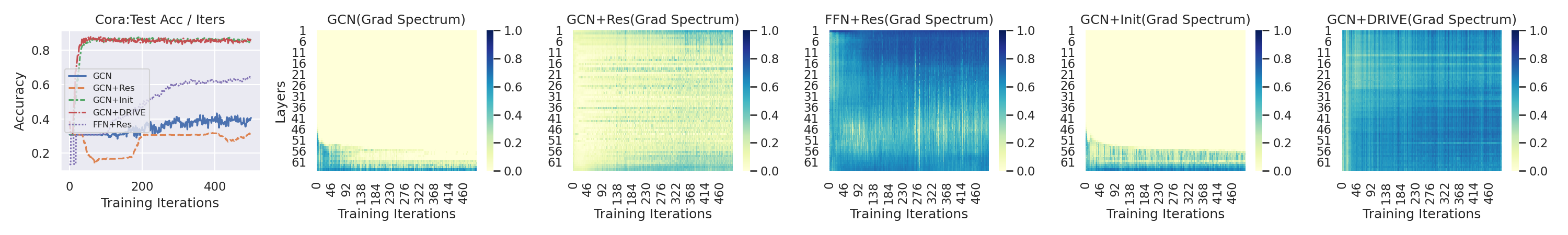}
         \caption*{(a) \textit{Cora(Homo)} dataset}
     \end{subfigure}
     \begin{subfigure}
         \centering
         \includegraphics[width=\textwidth]{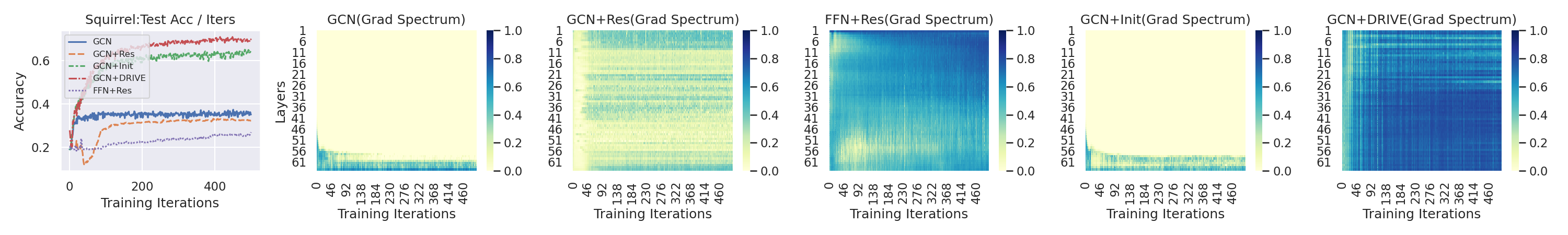}
         \caption*{(b) \textit{Squirrel(Hete)} dataset}
     \end{subfigure}
     \begin{subfigure}
         \centering
         \includegraphics[width=\textwidth]{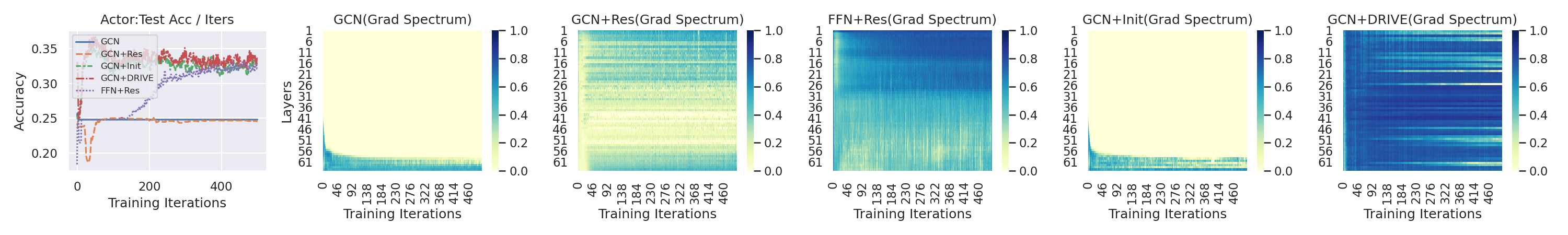}
         \caption*{(c) \textit{Actor(Hete)} dataset}
     \end{subfigure}
    \caption{The test accuracy and grad spectral property of different skip connection variants of 64 layers during training over both homophily and heterophily datasets.}
    \label{fig:three_grad_spectral}
\end{figure}

From Figure~\ref{fig:three_grad_spectral}, we have the following observations: 
(1) Due to stacking the propagation matrix P, the gradient of shallow layers for the GCN and GCN(Init) are over-smoothed over all datasets, e.g., the von Neumann entropy tends to 0. 
However, the performance of GCN(Init) is strongly better than that of GCN under 64 layers, which indicates that the GCN(Init) does not suffer the over-smoothed of features.
The reason is that the initial node features directly pass through to the deeper layer, and the gradient of the deeper layer is informative, which would adequately classify the nodes according to the initial features.
Therefore, the initial connection can make the behavior of the DeepGCN analogous to the shallow model and maintain performance. 
(2) For the GCN(Res), compared to the grad spectral information with the FFN(Res), the gradient of GCN(Res) is also over-smoothed due to the stacking of propagation matrix $P$. Moreover, compared to the performance with the GCN(Init), the binomial distribution of the longer path dominates the final representation. The over-smoothed features and over-smoothed gradient cause the identity mapping to be hard to optimize and performance degeneration when increasing layers.
(3) The gradient evolution of the DRIVE connection is as informative as the FFN(Res), indicating that the DRIVE connection can effectively solve the gradient smoothness problem. Therefore, GCN(DRIVE) achieves the best performance and converges quickly over all datasets.

All these results verify the backward path decomposition gradient analysis. Unlike the previous over-smoothing study, we show that the gradient smoothness problem is the key factor why the residual connection can not help DeepGNNs optimize to learn identity mapping properly. 

\section{More Experiments of UDGNNs}\label{supp:more_experiments}
In this section, we report more empirical results to show the
effectiveness of our Universal Deep GNN framework.


\subsection{Comparison of SOTA with different skip connections in terms of layer}

\begin{figure}[ht]
\centering
\centerline{\includegraphics[width=\linewidth]{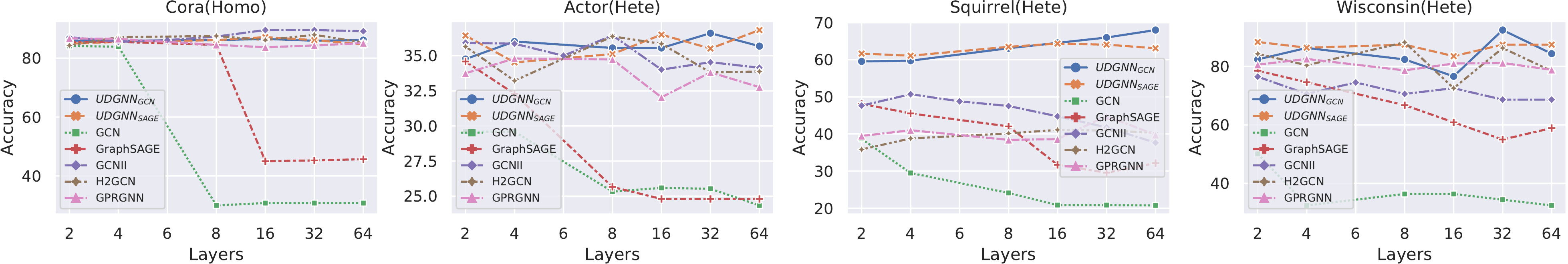}}
\caption{The performance comparison of the state-of-the-art DeepGNNs and our UDGNNs for different layers.
}
\label{fig:abalation_sota}
\end{figure}

We compare the performance between state-of-the-art DeepGNNs (GCNII, H2GCN, GPRGNN) and UDGNNs when increasing layers. 
\begin{itemize}
    \item GCNII~\cite{chen2020simple} combines the initial connection and identity mapping by explicitly modifying the graph convolution kernel to overcome over-smoothing and achieve state-of-the-art performance on homophily datasets. However, the constant weight parameter for initial connection and identity mapping would restricts the performance on heterophily datasets.
    \item H2GCN~\cite{zhu2020beyond} proposes three designs with separate ego and neighbors, high-order neighbors, and a combination of intermediate representations by JK connection.
    \item GPRGNN~\cite{chien2020adaptive} decouples the propagation and transformation in graph convolution and learns an arbitrary polynomial graph filter to incorporate multi-scale information by the adaptively generalized PageRank.
\end{itemize}
From Figure~\ref{fig:abalation_sota}, we have the following observations: (1) The performance of GCN and GraphSAGE drops rapidly as the number of layers grows. 
(2) GCNII, H2GCN, and GPRGNN are robust to the over-smoothing, especially on the homophily Cora dataset. However, their performance is not optimal in the heterophily scenario. The reason may be that the non-smooth heterophily may tend to over-smoothing even at shallow layers, which requires the model to adaptively control the smoothness.
(3) Compared with the other DeepGNNs, our UDGNN can help classical GNNs (GCN and GraphSAGE) become robust to the over-smoothing on both homophily and heterophily datasets. Moreover, due to the smoothness control ability, the UDGNN framework can well solve the heterophily datasets, e.g., increase the accuracy when stacking more layers on the Squirrel dataset.

\subsection{Effect of variants of normalization in residual connection}
To show the effectiveness and elegance of our DRIVE connection for preventing the over-smoothing, we replace the DRIVE connection in UDGNN* with the residual connection and different normalizations, including BatchNorm, LayerNorm, NodeNorm, and PairNorm.
\begin{align}
H^{l+1} &= \operatorname{Norm}(H^l+\operatorname{GraphConv}(A, H^l) ). 
\end{align}

\begin{figure}[ht]
\centering
\centerline{\includegraphics[width=\linewidth]{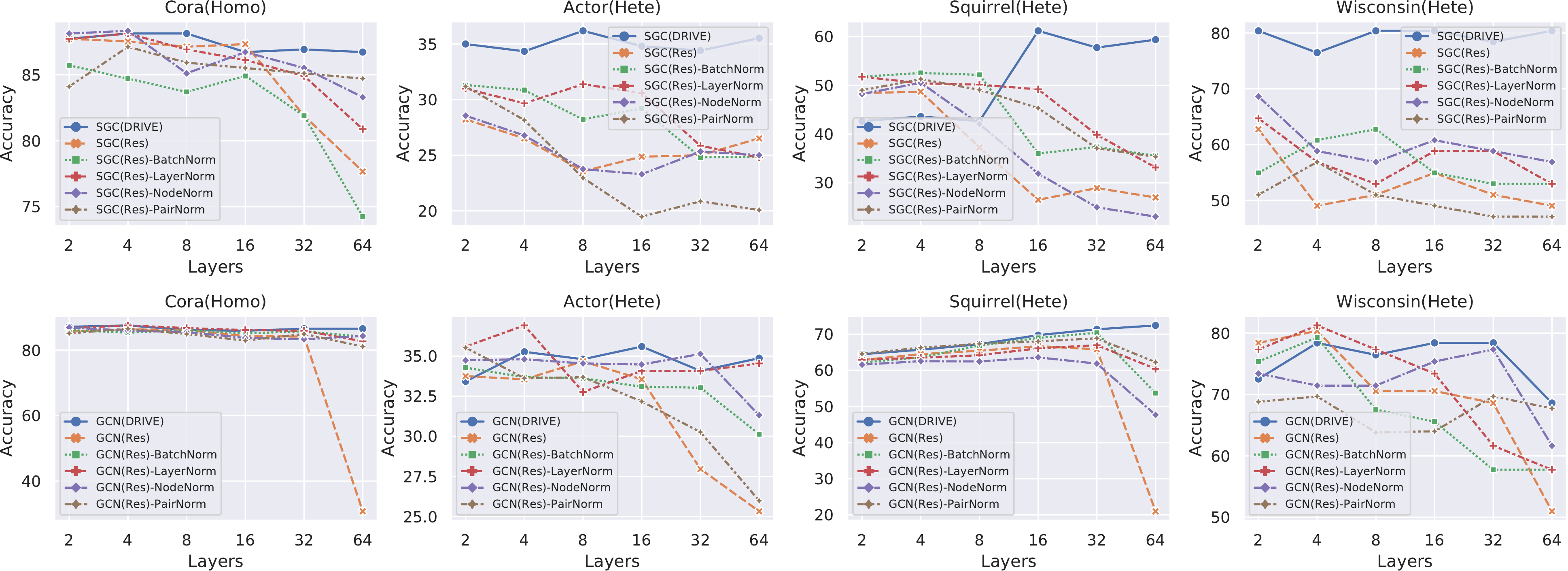}}
\caption{The effect of the recent normalization techniques for helping residual connection to prevent over-smoothing. Our DRIVE connection can prevent over-smoothing without any normalization.
}
\label{fig:abalation_norm}
\end{figure}

From Figure~\ref{fig:abalation_norm}, we can find that most normalization can slow down the performance degeneration of residual connections. However, their performance under the heterophily datasets is not optimal when we increase the number of layers. In contrast, without any normalization technique, our DRIVE connection can achieve the best performance over all datasets, especially for the SGC.

\subsection{Effect of the initialization of $\alpha$ in the DRIVE connection}

\begin{figure}[ht]
\centering
\centerline{\includegraphics[width=\linewidth]{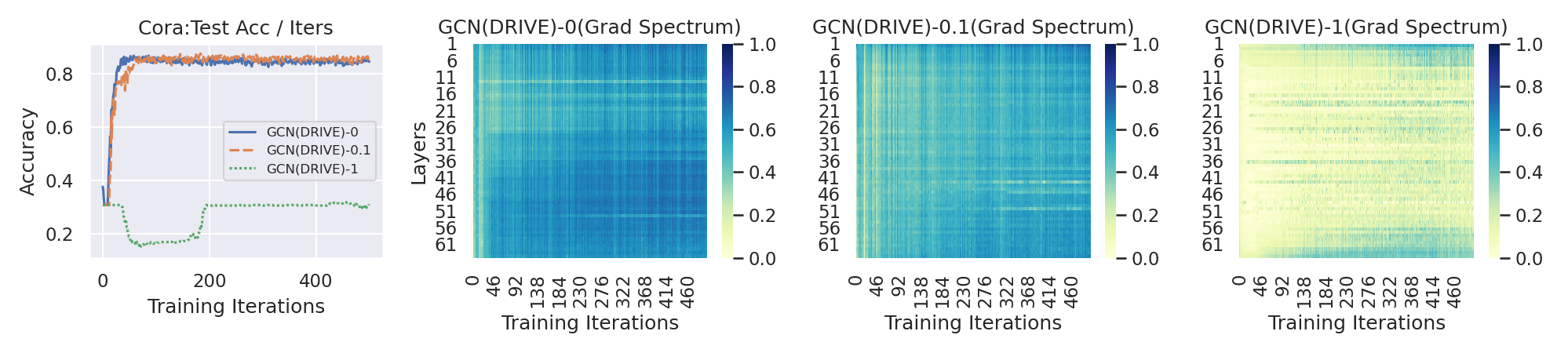}}
\caption{The test accuracy and grad spectral property of different initializations for $\alpha$ in the 64-layer GCN(DRIVE) during training over the Cora dataset.
}
\label{fig:alpha_grad}
\end{figure}

Although the parameter $\alpha$ is a learnable weight parameter in the DRIVE connection, to show the importance of the initialization of 0 to make the entire network behavior as a direct pass identity mapping, we compare the performance of different initializations of $\alpha$ in UDGNN*. 
\begin{figure}[ht]
\centering
\centerline{\includegraphics[width=\linewidth]{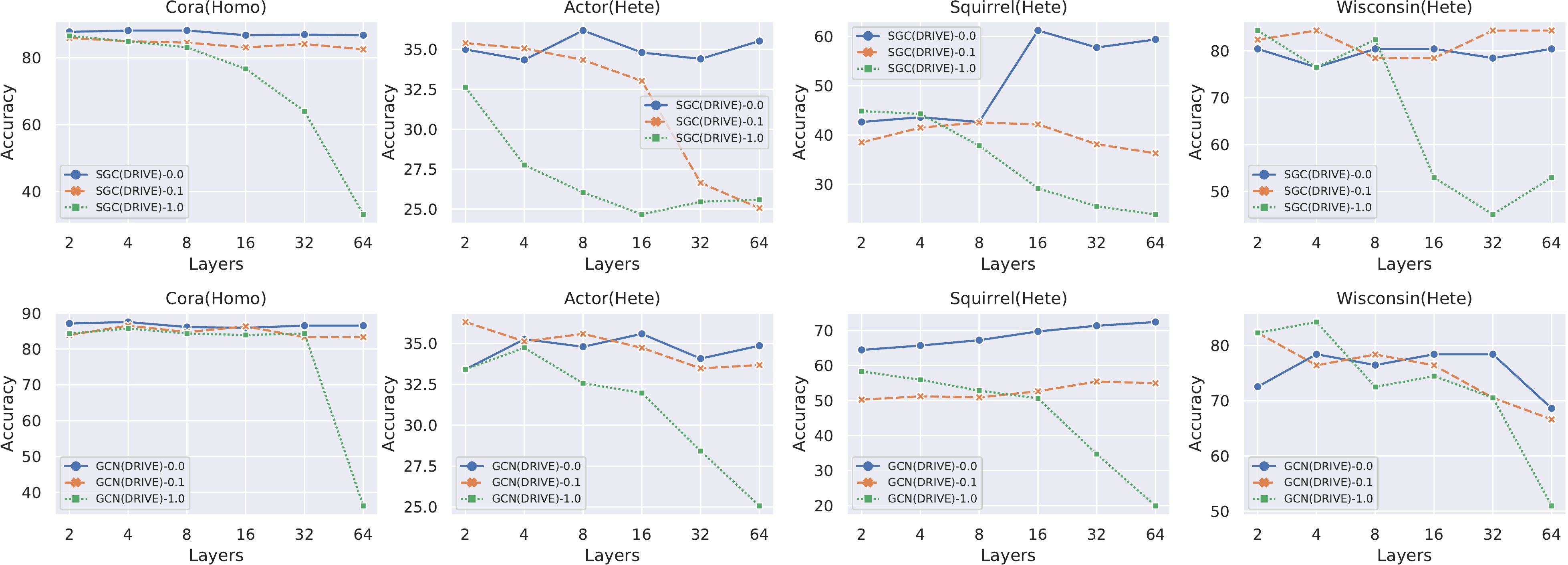}}
\caption{Ablation study on variants initialization of $\alpha$ for DRIVE connection in UDGNN* over both homophily and heterophily datasets.
}
\label{fig:abalation_alpha}
\end{figure}
From the results in Figure~\ref{fig:abalation_alpha}, we can observe that whether GraphConv is a GCN or SGC (with/without weight matrix $W$), the initialization of $\alpha$ is vital to the performance in terms of layers. For the initialization of 1, the beginning behavior of the model is similar to the residual connection. As shown in the right of Figure~\ref{fig:alpha_grad}, the GCN(DRIVE)-1 ($\alpha$ = 1, and layers = 64) would also suffer from the over-smoothed features and gradient, and cause optimization difficulty, although $\alpha$  is learnable. Therefore, the alpha failed to optimize and control smoothness properly, and the model's performance also degenerates when increasing the layers. On the one hand, compared with 1, the initialization of 0.1 can solve the over-smoothing to some extent, but the performance is still not optimal. On the other hand, the 0 initialization can achieve the best performance across all datasets. Hence, both the initialization of 0 and the learnable ability of $\alpha$ in DRIVE connection are vital for preventing over-smoothing.

\section{Limitation and Potential Negative Impact}\label{supp:limitation}
In this work, we proposed to rethink the over-smoothing problem of GNNs from a path decomposition view. One interesting finding is that we found the gradient smoothing phenomenon that prevents the residual connection in graph convolutions from optimizing to learn the identity mapping and to be anti the over-smoothing. Moreover, we proposed a DRIVE connection and a Universal Deep GNNs framework to make existing GNNs deeper and more powerful. Considering real-world graphs are complex and may be non-smooth, the Universal Deep GNN framework can easily unlock the potential of existing shallow GNNs and serve as a starting baseline for the architecture design of future GNNs.

Our analysis removes the non-linear ReLU activation for analysis. Although the ReLU may only slightly affect the performance according to the previous study, combining it into the total path decomposition analysis is interesting. Besides, we do not provide the convergence rate analysis of the over-smoothing. According to the~\cite{cong2021provable}, the convergence condition in terms of weight matrix $W$ and residual connection of the feature over-smoothing subspace theorem may also not be satisfied during training. We advocate peer researchers to investigate the over-smoothing problem of GNNs by considering the gradient optimization process. It would be interesting to analyze the convergence rate with respect to gradient smoothing.

\end{document}